\documentclass{article} 
\usepackage[preprint]{colm2026_conference}

\usepackage{microtype}
\usepackage{hyperref}
\usepackage{url}
\usepackage{booktabs}
\usepackage{graphicx} 
\usepackage{subcaption}
\usepackage{listings}
\usepackage{amsmath}
\usepackage{multirow}
\usepackage{enumitem}
\usepackage{adjustbox}
\usepackage{xcolor} 
\usepackage{tcolorbox}

\usepackage{lineno}

\definecolor{darkblue}{rgb}{0, 0, 0.5}
\hypersetup{colorlinks=true, citecolor=darkblue, linkcolor=darkblue, urlcolor=darkblue}

\definecolor{c}{cmyk}{0.2,0,0.8,0} 
\newtcolorbox{passage}{on line,colback=c!10,colframe=white,size=fbox,arc=3pt, box align=base, top=-2pt, bottom=0pt, boxrule=0pt, before=\adjustbox{valign=c}\bgroup, after=\egroup, before upper=\strut}

\title{In-Context Learning in Speech Language Models: Analyzing the Role of Acoustic Features, Linguistic Structure, and Induction Heads}

\author{
Charlotte Pouw\textsuperscript{1},
Hosein Mohebbi\textsuperscript{2},
Afra Alishahi\textsuperscript{2},
Willem Zuidema\textsuperscript{1} \\
\textsuperscript{1}ILLC, University of Amsterdam \quad
\textsuperscript{2}CSAI, Tilburg University \\
\texttt{\{c.m.pouw,w.h.zuidema\}@uva.nl} \quad
\texttt{\{h.mohebbi,a.alishahi\}@tilburguniversity.edu}
}

\begin{document}

\ifcolmsubmission
\linenumbers
\fi

\maketitle

\begin{abstract}

In-Context Learning (ICL) has been extensively studied in text-only Language Models, but remains largely unexplored in the speech domain. Here, we investigate how linguistic and acoustic features affect ICL in Speech Language Models. We focus on the Text-to-Speech (TTS) task, which allows us to analyze ICL from two angles: (1) how accurately the model infers the task from the demonstrations (i.e., generating the correct spoken content), and (2) to what extent the model mimics the acoustic characteristics of the demonstration speech in its output. We find that speaking rate strongly affects ICL performance and is also mimicked in the output, whereas pitch range and intensity have little impact on performance and are not consistently reproduced. Finally, we investigate the role of induction heads in speech-based ICL and show that these heads play a causal role: ablating the top‑\textit{k} induction heads completely removes the model’s ICL ability, mirroring findings from text-based ICL.

\end{abstract}

\section{Introduction}

In-Context Learning (ICL) is one of the key abilities contributing to the success of autoregressive Large Language Models (LLMs), which allows them to perform multiple downstream tasks based on a few in-context demonstrations provided in the input, without updating the model parameters \citep{NEURIPS2020_1457c0d6}. While ICL has been extensively studied in the text domain \citep{dong2024survey}, its application to speech remains relatively underexplored.
The emergence of Speech Language Models (SpeechLMs) has started to close this gap by unifying audio and text within a single modeling framework. These models jointly model speech and text as sequences of discrete tokens and are trained via next-token prediction, analogous to how LLMs model text tokens \citep{cui-etal-2025-recent}, allowing them to flexibly perform speech-only, text-only, and speech–text tasks (such as Automatic Speech Recognition (ASR) and Text-to-Speech (TTS)) through ICL \citep{nguyen-etal-2025-spirit}.

In text-only LLMs, it has been shown that non-trivial factors affect ICL performance. For classification tasks, prompt formatting and input distribution (i.e., whether examples are in- or out-of-domain) matter more than label correctness \citep{min-etal-2022-rethinking}, and ICL only emerges for specific architectures and training data distributions \citep{chan2022data}. Understanding \textit{why} these factors matter has motivated mechanistic interpretability research, which seeks to uncover the functional circuits that implement ICL within the network \citep{Ferrando2024APO, wiegreffe2024mechanistic}. This line of work has found that \textit{induction heads} (attention heads implementing a \emph{copying} function \citep{elhage2021mathematical}) appear central to ICL \citep{olsson2022context}, and ablating them significantly degrades performance on both NLP and abstract pattern recognition tasks \citep{crosbie2025induction}. Whether similar mechanisms play a role in speech-based ICL, where models must reason over both acoustic and linguistic structure, remains an open question.

While both ASR and TTS require SpeechLMs to learn a mapping between the speech and text in the demonstrations, they differ in output modality. In ASR, the output is text, so ICL performance can only be measured in terms of transcription accuracy. In TTS, the output is speech, which means that ICL can be measured in two ways: (1) how well the model infers the task (i.e., how well it produces spoken content that matches the textual content of the target), and (2) whether the model additionally \textit{mimics} the acoustic properties of the demonstration speech in its output (e.g., fast speaking rate in the demonstration speech yielding fast speaking rate in the output). This second dimension makes TTS uniquely suited for evaluating speech-based ICL, investigating how SpeechLMs exploit both the linguistic and acoustic features within the in-context demonstrations.

In this work, we systematically isolate the effect of various acoustic and linguistic features on ICL performance for TTS, and quantify to what extent SpeechLMs learn to mimic the acoustic features of demonstration speech. We additionally investigate whether induction heads play a role in speech-based ICL, as they do in text-based ICL. To our knowledge, this is the first study to examine ICL mechanisms in SpeechLMs for TTS specifically.


\section{Related Work}

We discuss related work on the role of demonstrations in speech-based ICL, as well as mechanistic studies of ICL in the text domain.

\subsection{What makes good examples for speech-based ICL?}

Numerous studies have shown that the choice of demonstrations strongly affects ICL performance, both in text models \citep[e.g.,][]{zhao2021calibrate} and in vision models \citep{zhang2023makes}, and that demonstrations that are semantically similar to the target improve performance \citep[e.g.,][]{liu2022makes}. In speech-based ICL (ASR specifically), studies have found that demonstrations matching the speaker, language variety, or semantic content of the target improve ASR performance compared to non-matching demonstrations \citep{wang2024can,roll2025context,linke2025context}. However, these studies did not explicitly study the effect of acoustic and linguistic features in isolation. We extend this line of work by explicitly disentangling the role of individual acoustic and linguistic features on ICL performance in TTS and, importantly, by analyzing whether ICL can emerge when there is no linguistic overlap at all between the demonstration and target.




\subsection{Mechanism behind ICL}\label{sec:related-work-mechanism-icl}
Several accounts have been proposed for the mechanism underlying ICL. One line of work characterizes it as an implicit form of gradient descent \citep{von2023transformers}, while another frames it as error-driven learning \citep{Chang2006BecomingS}, linking it to the \textit{inverse frequency effect} observed in humans, where less frequent items have a stronger influence on behavior than more frequent ones \citep{ jumelet2024language, zhou2025context}. A complementary perspective comes from mechanistic interpretability. \citet{elhage2021mathematical} identified \textit{induction heads}: attention heads that locate prior occurrences of the current token and copy what followed them. \citet{olsson2022context} argue that these heads are the primary mechanism underlying ICL in Transformers, showing that their emergence during training coincides with a sharp improvement in in-context learning ability. This finding was later corroborated by \cite{crosbie2025induction}, who show that ablating the top-\textit{k} induction heads significantly degrades ICL performance on a range of NLP and abstract pattern recognition tasks. To our knowledge, we are the first to analyze the role of such heads in speech-based ICL.


\section{Experiments}

\begin{figure}[t]
\centering 
\includegraphics[width=0.8\textwidth]{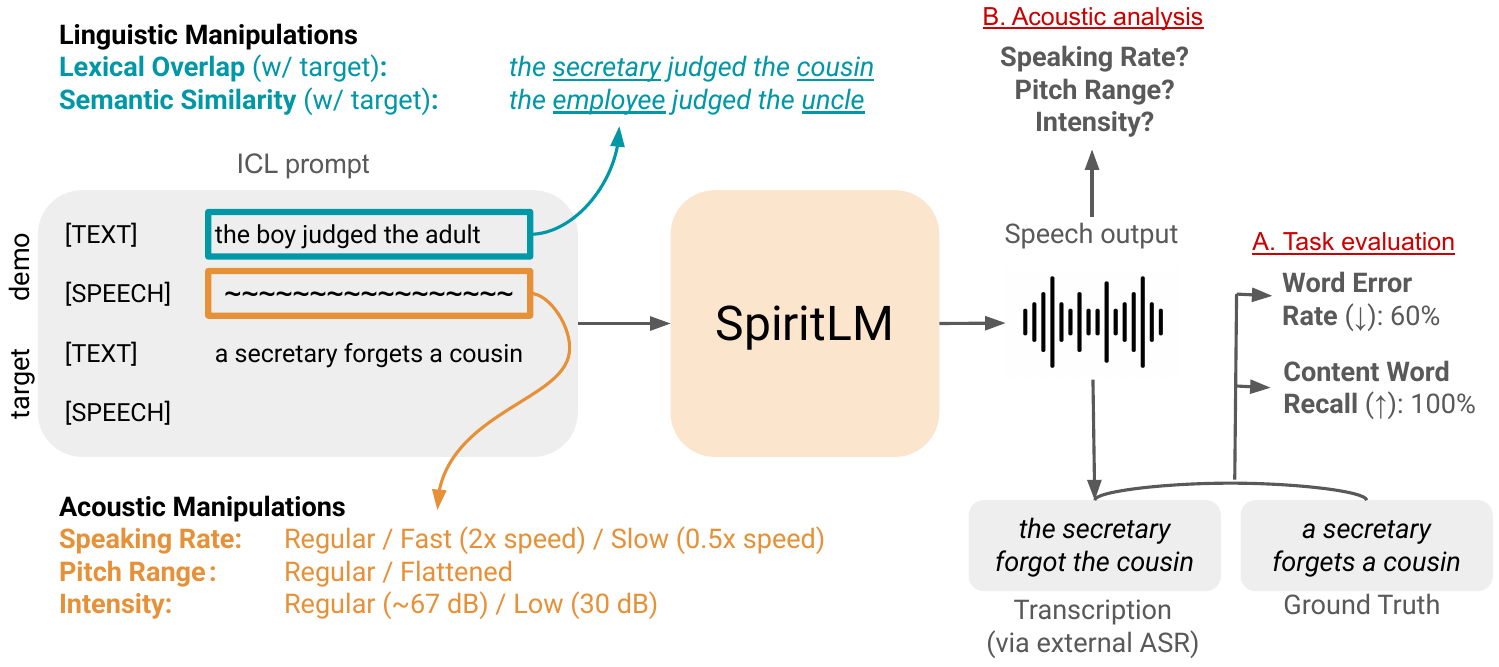}
\caption{Our experimental setup: we construct ICL prompts and apply different linguistic and acoustic manipulations. We then (A) evaluate SpiritLM's performance on the TTS task under each manipulation, and (B) measure the acoustic characteristics of the output speech.}
\label{fig:method}
\end{figure}

\subsection{Speech Language Model}

We investigate \textsc{SpiritLM} \citep{nguyen-etal-2025-spirit}, an adapted version of \textsc{Llama-2-7B} \citep{touvron2023llama} continuously pretrained on interleaved sequences of speech and text tokens. The architecture consists of three components: a \textsc{HuBERT} model \citep{hsu2021hubert} that encodes raw audio into discrete speech units, a Llama-2-7B language model that performs next-token prediction over the combined speech--text vocabulary, and a \textsc{HiFi-GAN} vocoder \citep{kong2020hifi} that synthesises a waveform from the predicted speech tokens. As we use \textsc{SpiritLM} for TTS, we fix the output modality to \emph{speech} throughout all experiments. Configuration details are listed in Appendix \autoref{appendix-sec:generation-setup}.

\subsection{Data}\label{sec:data}

\paragraph{Corpus.}
We use the \textsc{Prime-LM} corpus from \citet{sinclair2022structural}, which contains controlled pairs of demonstration sentences (i.e., \textit{primes}) and corresponding target sentences. The corpus includes a baseline condition called \textit{Core}, in which demonstrations and targets are explicitly designed to avoid any linguistic overlap: they share no lexical items, exhibit no semantic similarity, and differ in verb tense and function words. This condition serves as the basis throughout our experiments. 

\paragraph{Speech Synthesis.} To construct demonstrations for the TTS task, we need both a textual and a spoken version of each sentence. Hence, we synthesize all sentences using two different voices: (1) an American English female voice from SpeechT5\footnote{\url{https://huggingface.co/microsoft/speecht5_tts}} \citep{ao2022speecht5}\footnote{voice = ``clb'' from CMU Arctic x-vectors; \url{https://huggingface.co/datasets/Matthijs/cmu-arctic-xvectors}}; (2) an American English female voice from Kokoro-TTS\footnote{voice = ``af\_heart''; \url{https://huggingface.co/hexgrad/Kokoro-82M}}. This results in two text–speech pairs per sentence, allowing us to evaluate the robustness of our results across speakers.

\paragraph{Factors under Investigation.} We analyze how ICL performance on the \textit{Core} condition is affected by four factors, each described below.

\begin{enumerate}[leftmargin=*, label=\textbf{(\arabic*)}]
    \item \textbf{Number of Demonstrations.} The \textit{Core} condition includes a \textit{Cumulative} variant in which each target is paired with 1--5 demonstrations, allowing us to analyze whether more demonstrations lead to better ICL performance.
    \item \textbf{Syntactic Structure.} Both the demonstration and the target sentences in \textsc{Prime-LM} are available in multiple syntactic structures. We specifically use transitive sentences in active and passive voice (e.g., active: \textit{the boy judged the adult}; passive: \textit{the adult was judged by the boy)}. This yields four combinations: (1) active targets paired with active demonstrations (congruent syntax); (2) active targets paired with passive demonstrations (incongruent syntax); (3) passive targets paired with passive demonstrations (congruent syntax); (4) passive targets paired with active demonstrations (incongruent syntax). These combinations allow us to analyze whether syntactic congruence between demonstration and target matters for ICL performance.
    \item \textbf{Linguistic Overlap.} Building on the \textit{Core} baseline, \textsc{Prime-LM} includes additional conditions that introduce controlled overlap between demonstration and target, allowing us to isolate the contribution of individual linguistic features. We consider two types of overlap: \textit{Lexical Overlap} and \textit{Semantic Similarity}, each available in different degrees (i.e., \textit{nouns only}, \textit{main verb only}, and for Lexical Overlap, also \textit{function words only} and \textit{all words}). Illustrative examples are shown in \autoref{tab:conditions}. All target sentences in the linguistic overlap condition are paired with a single demonstration only.
    \item \textbf{Acoustic Manipulations.} We apply controlled acoustic manipulations to the \textit{Core} speech to examine how acoustic features affect ICL performance, independent of linguistic overlap. Specifically, we manipulate \textbf{speaking rate} (\textit{Fast Speaking Rate} = 2x original speed, \textit{Slow Speaking Rate} = 0.5x original speed), \textbf{pitch range} (\textit{Flattened Pitch} = pitch flattened to the mean of each individual utterance), and \textbf{intensity} (\textit{Low Intensity} = intensity reduced from $\sim67\,\text{dB}$ to 30 dB; we do not increase intensity as this often leads to clipping). While not comprehensive, this selection of manipulations is designed to cover the primary acoustic dimensions of time, frequency, and amplitude. We use Praat \citep{praat} for all acoustic manipulations.\footnote{We use Praat scripts by Matthew B. Winn for pitch manipulation and intensity scaling \url{http://www.mattwinn.com/praat.html}} Because we derive all acoustic variants from the \textit{Core} condition, we can assess how acoustic manipulations interact with the number of demonstrations (1-5).
\end{enumerate}

For each condition, we sample 100 target sentences, each paired with 10 different demonstration sets. For the linguistic overlap conditions, each demonstration set consists of a single demonstration, while for the acoustic manipulation conditions, each set consists of 1--5 demonstrations. All conditions are crossed with four syntactic combinations (active/passive target
× congruent/incongruent demonstration), yielding 4,000 items per linguistic overlap condition and 20,000 items per acoustic manipulation condition.

\begin{table}[t]
\centering
\resizebox{0.8\textwidth}{!}{%
\begin{tabular}{llll}
\toprule
\textbf{Condition} & \textbf{Degree of Overlap} & \textbf{Demonstration} & \textbf{Target} \\
\midrule
\textit{Core}
    & ---
    & \textit{the boy judged the adult}
    & \textit{a secretary forgets a cousin} \\
\midrule
\multirow{4}{*}{\textit{Lexical Overlap}}
    & nouns only
    & \textit{the \underline{secretary} judged the \underline{cousin}}
    & \multirow{4}{*}{\textit{a secretary forgets a cousin}} \\
  & main verb only
    & \textit{the boy \underline{forgot} the adult}
    & \\
  & function words only
    & \textit{\underline{a} boy judges \underline{an} adult}
    & \\
  & all words
    & \textit{\underline{a secretary forgets a cousin}}
    & \\
\midrule
\multirow{2}{*}{\textit{Semantic Similarity}}
    & nouns only
    & \textit{the \underline{employee} judged the \underline{uncle}}
    & \multirow{2}{*}{\textit{a secretary forgets a cousin}} \\
  & main verb only
    & \textit{the boy \underline{forgave} the adult}
    & \\
\bottomrule
\end{tabular}%
}
\caption{Illustrative examples of linguistic overlap conditions. Examples are shown for sentences with \textit{active} syntax; the same conditions apply to the \textit{passive} sentences. 
The overlapping linguistic elements between the demonstration and target sentences are \underline{underlined}. Note that verb tense deliberately only agrees between demonstration and target in the \textit{function words only} and \textit{all words} condition.
}
\label{tab:conditions}
\end{table}

\subsection{ICL Prompt Structure}

Following \cite{nguyen-etal-2025-spirit}, we prompt SpiritLM to perform TTS using alternating sequences of   \texttt{[TEXT]} and \texttt{[SPEECH]} tokens, separated by special ``stop" tokens. For example, given the demonstration sentence \textit{the boy judged the adult} and the target sentence \textit{a secretary forgets a cousin} (see \autoref{fig:method}), the ICL prompt is constructed as follows:

\begin{passage}
\begin{lstlisting}[basicstyle=\ttfamily\small]
[TEXT] the boy judged the adult `stop'
[SPEECH] <demo speech> <speech:STOP>

[TEXT] a secretary forgets a cousin `stop'
[SPEECH]
\end{lstlisting}
\end{passage}



Here, \texttt{<demo speech>} represents the speech token sequence for \textit{the boy judged the adult}, and \texttt{<speech:STOP>} represents the speech token sequence for the spoken utterance “stop”.  
In the zero-shot setting (i.e., no demonstrations), only the second text-speech block is provided.  
In the one-shot setting, the first block serves as the demonstration. The prompt can then be extended to few-shot by including multiple demonstration blocks.

\subsection{Evaluation}

\subsubsection{Task Performance}
To assess how well SpiritLM performs the TTS task, we transcribe all speech outputs using an external ASR model, \texttt{whisper-large-v3-turbo} \citep{radford2022whisper}\footnote{\url{https://huggingface.co/openai/whisper-large-v3-turbo}. We find that this version of Whisper performs most consistently on different speech variations, see Appendix \autoref{appendix-sec:whisper-size-comparison}.}, and measure Word Error Rate as the standard metric for evaluating transcription quality in ASR. In our experiments, we observed that SpiritLM often predicts the wrong function words or verb inflections, but still predicts the correct content words. Hence, to capture this, we also define and report \textbf{Content Word Recall}, where we measure the percentage of correctly predicted content words (lemmatized). Before computing these metrics, we clean the transcriptions by removing punctuation, removing the special token ``stop'', and lowercasing all words.

\subsubsection{Acoustic Mimicking}
To evaluate the extent to which SpiritLM acoustically mimics the demonstration speech, we measure, for each output, the acoustic features that were explicitly manipulated in the corresponding demonstration. Specifically, for outputs that followed demonstrations with \textit{Fast Speaking Rate} and \textit{Slow Speaking Rate}, we measure speaking rate; for outputs that followed demonstrations with \textit{Flattened Pitch}, we compute pitch statistics; and for outputs that followed demonstrations with \textit{Low Intensity (30 dB)}, we compute intensity statistics. As a baseline, we compute all acoustic features of outputs that followed \textit{Core} demonstrations, and compare all other conditions against this baseline. We compute speaking rate (in words per second) by dividing the number of words in Whisper's transcription by the duration of the utterance. We use Praat to extract pitch and intensity statistics (mean, minimum, and maximum over the whole utterance).

\section{Results}

We present the results by first evaluating the performance of SpiritLM on the TTS task under each manipulation of the demonstrations, followed by an analysis of the extent to which the acoustic characteristics of the demonstrations are mimicked in the target output. Since we observe highly similar patterns across the two synthetic voices used to generate the spoken prompts, we report the Kokoro-TTS results in the main text and relegate the SpeechT5 results to Appendix \autoref{appendix-sec:speecht5-results}.

\subsection{Performance on the TTS task}

\begin{figure}[t]
\centering 
\includegraphics[width=0.8\textwidth]{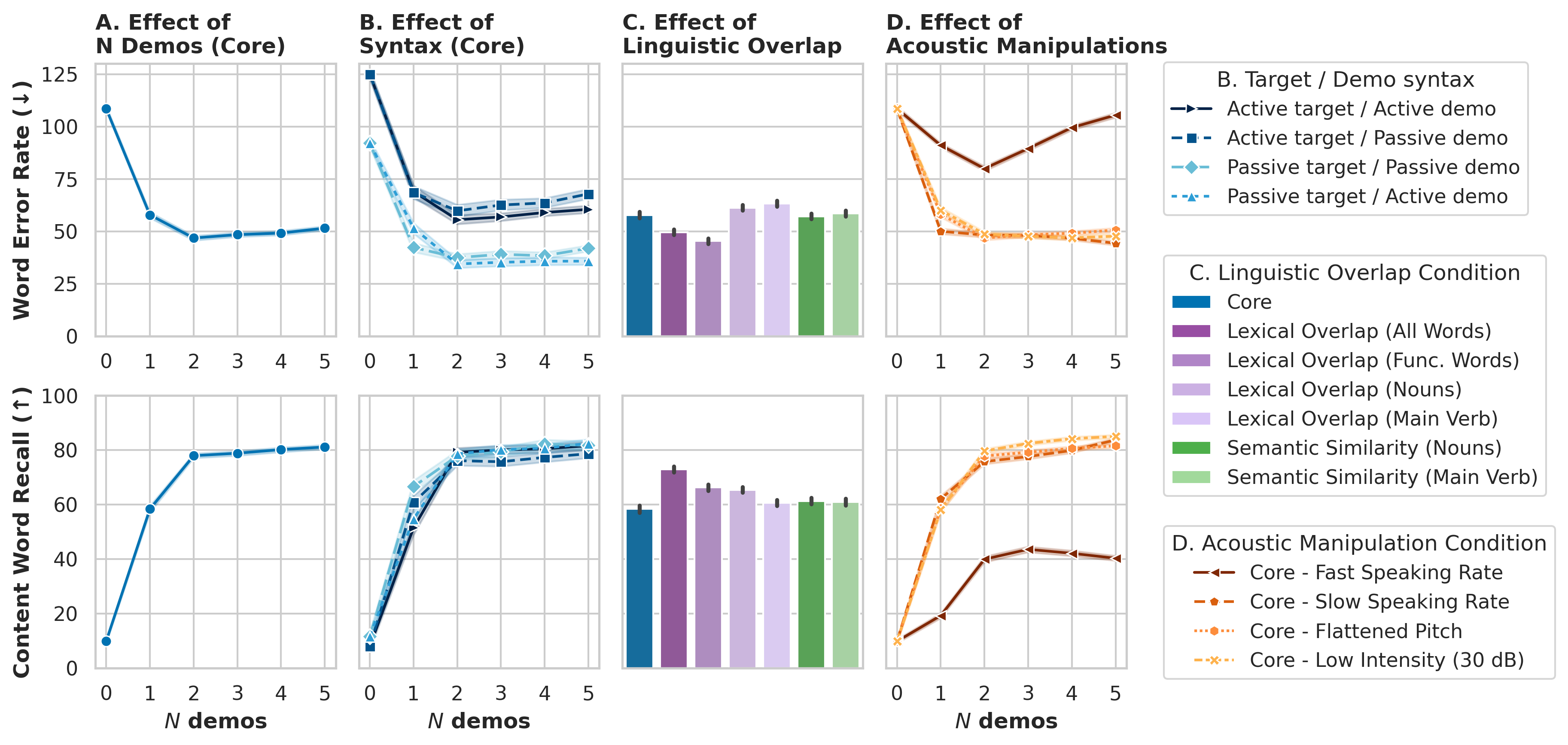}
\caption{ICL performance of SpiritLM given demonstrations synthesized by KokoroTTS. \textbf{(A)}: performance on the \textit{Core} condition across 1-5 demonstrations; \textbf{(B)}: performance on the \textit{Core} condition, separated by the syntactic structure of the target and the demonstration; \textbf{(C)}: performance under various types of linguistic overlap between demonstration and target; \textbf{(D)}: performance under various acoustic manipulations of the demonstrations. Shaded regions/error bars indicate 95\% confidence intervals.}
\label{fig:ling_manip_kokoro}
\end{figure}

\subsubsection{Observation 1: A single demonstration is sufficient to trigger ICL}

In \autoref{fig:ling_manip_kokoro}A, we see that the 1-demo condition clearly outperforms the 0-demo baseline, indicating that SpiritLM learns the TTS task given merely a single demonstration. We observe that the performance improves further when the model receives 2 demonstrations, but after that, the performance plateaus. The fact that we see ICL behavior in the \textit{Core} condition indicates that linguistic overlap between demonstration and target is not strictly necessary for ICL to emerge (although, as we discuss below, such overlap can boost performance).

\subsubsection{Observation 2: ICL performance depends on syntax}

\autoref{fig:ling_manip_kokoro}B shows that, under the Word Error Rate metric, the syntactic structure of the target is more important for ICL performance than the syntactic alignment between demonstration and target. Specifically, performance is consistently higher for passive targets than for active targets, and this advantage persists regardless of whether the demonstrations are syntactically congruent with the target. This pattern does not hold when evaluated using the Content Word Recall metric, which implies that the advantage for passive targets primarily stems from more accurate generation of function words (e.g., prepositions such as ``by"), rather than from differences in the realization of content words.

\subsubsection{Observation 3: Lexical overlap boosts ICL performance}

\autoref{fig:ling_manip_kokoro}C shows that \textit{Lexical Overlap} improves ICL performance relative to \textit{Core} for both metrics, while \textit{Semantic Similarity} does not. This suggests that semantic similarity between demonstration and target is less important for speech-based ICL than has previously been reported for text-based ICL \citep{liu2022makes}. When considering specific degrees of lexical overlap, the two evaluation metrics show slightly different patterns: for Word Error Rate, overlap in All Words and Function Words improves ICL performance, likely because shared determiners between demonstrations and targets directly reduce errors. For Content Word Recall, we additionally find that overlap in nouns further boosts ICL performance.

\subsubsection{Observation 4: Speaking rate affects ICL performance}

\autoref{fig:ling_manip_kokoro}D shows that most acoustic manipulations do not affect ICL performance, with one clear exception: demonstrations with a \textit{Fast Speaking Rate} lead to substantially worse ICL performance compared to \textit{Core}.\footnote{Note that this difference cannot be attributed to the transcription quality of Whisper, as we deliberately chose a version of Whisper that performs equally well across several speaking rates, see Appendix \autoref{appendix-sec:whisper-size-comparison}.} Demonstrations with a \textit{Slow Speaking Rate} do not have the same effect; in some cases (i.e., for 1 demonstration), we even observe that demonstrations with a \textit{Slow Speaking Rate} slightly improve performance relative to \textit{Core}. This indicates that with a \textit{Slow Speaking Rate}, the demonstrations contain more spread-out content for the model to learn from in context, whereas a \textit{Fast Speaking Rate} compresses the content and thus reduces the amount of useful in-context information.

\subsection{Mimicking of Acoustic Characteristics}

\begin{figure}[t]
\centering 
\includegraphics[width=0.9\textwidth]{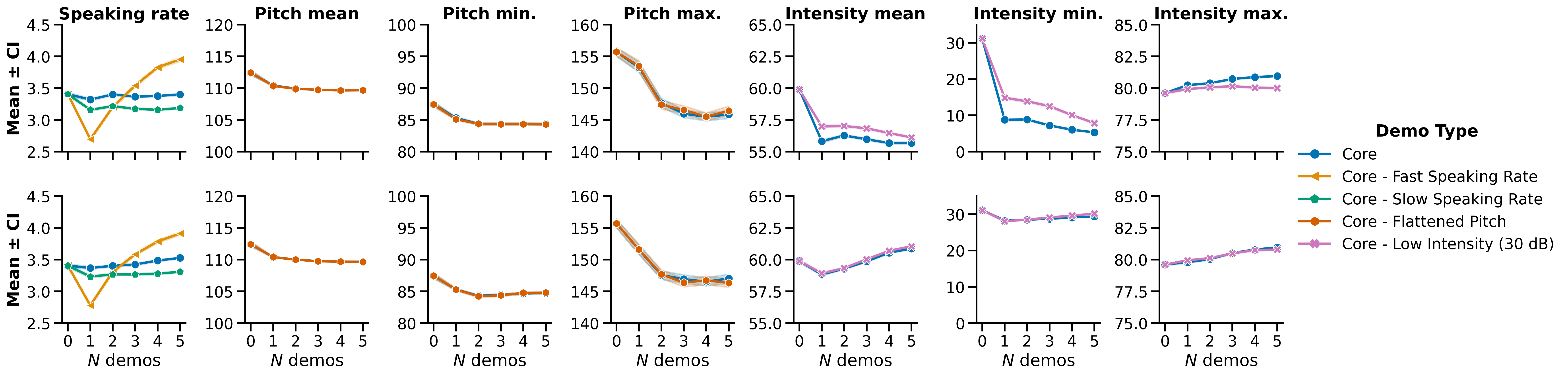}
\caption{Acoustic features of SpiritLM speech outputs given different demonstration types. Top row: demonstrations are synthesized using KokoroTTS; Bottom row: demonstrations are synthesized using SpeechT5. Shaded regions indicate 95\% confidence intervals.}
\label{fig:acoustic-mimicking}
\end{figure}

\autoref{fig:acoustic-mimicking} shows the acoustic features of the SpiritLM outputs under different acoustic manipulations of the demonstrations. We observe that speaking rate follows the expected pattern to some extent: after \textit{Fast Speaking Rate} demonstrations, outputs are faster than after \textit{Core} demonstrations, and after \textit{Slow Speaking Rate} demonstrations, outputs are slower. However, the \textit{Fast} effect only becomes clear after three demonstrations, whereas the \textit{Slow} effect is already visible after a single demonstration.

By contrast, the pitch range of the demonstrations (regular vs.\ flattened) does not affect the pitch range of the outputs. We do see a slight effect of the number of demonstrations: the maximum pitch decreases slightly with more demonstrations, which we attribute to the relatively flat prosody of the read-out sentences (e.g., \textit{the boy judged the adult}) compared to more expressive continuations in the 0-demo setting (verified by qualitative listening).

Finally, the intensity of the demonstrations only influences output intensity when the demonstrations are synthesized using KokoroTTS. With SpeechT5 demonstrations, we see no differences between \textit{Core} and \textit{Core - Low Intensity (30 dB)}. With KokoroTTS demonstrations, the mean intensity is slightly higher under the \textit{Low Intensity} condition than under \textit{Core}, with a slightly higher minimum and slightly lower maximum. This suggests a modest flattening of the intensity range after low-intensity demonstrations.

Overall, these results reveal a clear asymmetry across acoustic features: temporal features such as speaking rate, which directly affect the number of acoustic tokens, modulate both ICL performance and the speaking style of SpiritLM’s outputs. By contrast, features in the frequency and amplitude domains (pitch range and intensity) show little to no effect on ICL performance and only weakly influence the acoustic realization of the output.

\section{The Role of Induction Heads}

Induction heads have been shown to support ICL in text-only LLMs by implementing a prefix-matching and copying mechanism (see \autoref{sec:related-work-mechanism-icl}). In this section, we ask whether such heads are also involved in speech-based ICL, and whether there are modality-specific induction heads that operate on speech or text tokens specifically.


\subsection{Experiments}


\subsubsection{Identifying Induction Heads}

While induction heads should in principle be identified mechanistically by showing that the head exhibits both prefix-matching and copying, prior work suggests that induction heads can be reliably identified by computing ``prefix-matching scores'' as a behavioral proxy. Here, we consider two ways of computing prefix-matching scores: (1) the method used in previous work \citep{olsson2022context,bansal-etal-2023-rethinking,crosbie2025induction} which uses sequences of random tokens; and (2) an adapted version of this method that computes modality-specific prefix-matching scores within well-formed inputs (our ICL prompts).

We find that both methods yield a similar distribution of prefix-matching scores across heads, indicating that heads identified as strong prefix matchers on random sequences behave similarly on well-formed inputs. We report and analyze the ICL-based scores in the main text, and include the results from the random-sequence method in the Appendix (\autoref{appendix-sec:prefix-matching-random}).


\paragraph{Prefix-Matching on ICL Prompts.}\label{sec:pfx-matching-icl-prompts} Our adapted method for computing modality-specific prefix-matching scores is as follows: for each token $t$ at position $i$, we check whether $t_i$ occurred earlier in the sequence (we refer to this token as $t_j$ at position $j < i$). If such a $t_j$ exists, (1) we include the attention from $t_i$ to $t_{j+1}$ (i.e., the token immediately following $t_j$) in the \emph{prefix-matching} attention sum; and (2) as a baseline, we include the attention from $t$ to $t_r$ (where $t_r$ is a randomly chosen token in the previous context, with $r < i$ and $r \neq j+1$), in the \emph{non-prefix-matching} attention sum. We compute these sums separately for speech and text tokens, and then normalize by the number of speech or text tokens, respectively, to obtain average prefix-matching and non-prefix-matching scores per head and per modality. We repeat this procedure across 40 ICL prompts (1 target sentence $\times$ 10 demonstration sets $\times$ 4 syntax combinations) and for three conditions (\textit{Core}, \textit{Core -- Fast Speaking Rate}, \textit{Core -- Slow Speaking Rate}). Thus, each prefix-matching score for each head is averaged over 120 ICL sequences in total. We use prompts from the 2-demonstration setting, as we observe the best performance there for all three conditions.

\paragraph{Head Groups.}\label{subsec:head-groups} Using these results, we define six subsets of heads (where we choose $k$=50): \textbf{(1) Speech-Prefix Heads (Top):} the top-$k$ heads with the highest prefix-matching scores on speech tokens; \textbf{(2) Text-Prefix Heads (Top):} the top-$k$ heads with the highest prefix-matching scores on text tokens; \textbf{(3) Speech-Prefix Heads (Unique):} the heads in the Speech-Prefix Heads (Top) set that are not in the Text-Prefix Heads (Top) set. \textbf{(4) Text-Prefix Heads (Unique):} the heads in the Text-Prefix Heads (Top) set that are not in the Speech-Prefix Heads (Top) set; \textbf{(5) Non-Prefix Heads (Top):} the top-$k$ heads with the highest non-prefix-matching scores across the two modalities; and \textbf{(6) Random Heads:} a random selection of $k$ heads from the remaining heads that are not in any of the aforementioned groups.

\subsubsection{Ablating Induction Heads}

For each head group, we evaluate its causal role in ICL by ablating all heads in the group and re-computing ICL performance on the full set of ICL prompts (4000 for each condition). We compare the resulting performance to the ICL performance of the full model (without any ablations). We ablate heads by setting their output vectors to zero.

\begin{figure}[t]
    \centering
    
    \begin{subfigure}{0.8\textwidth}
        \centering
        \includegraphics[width=\textwidth]{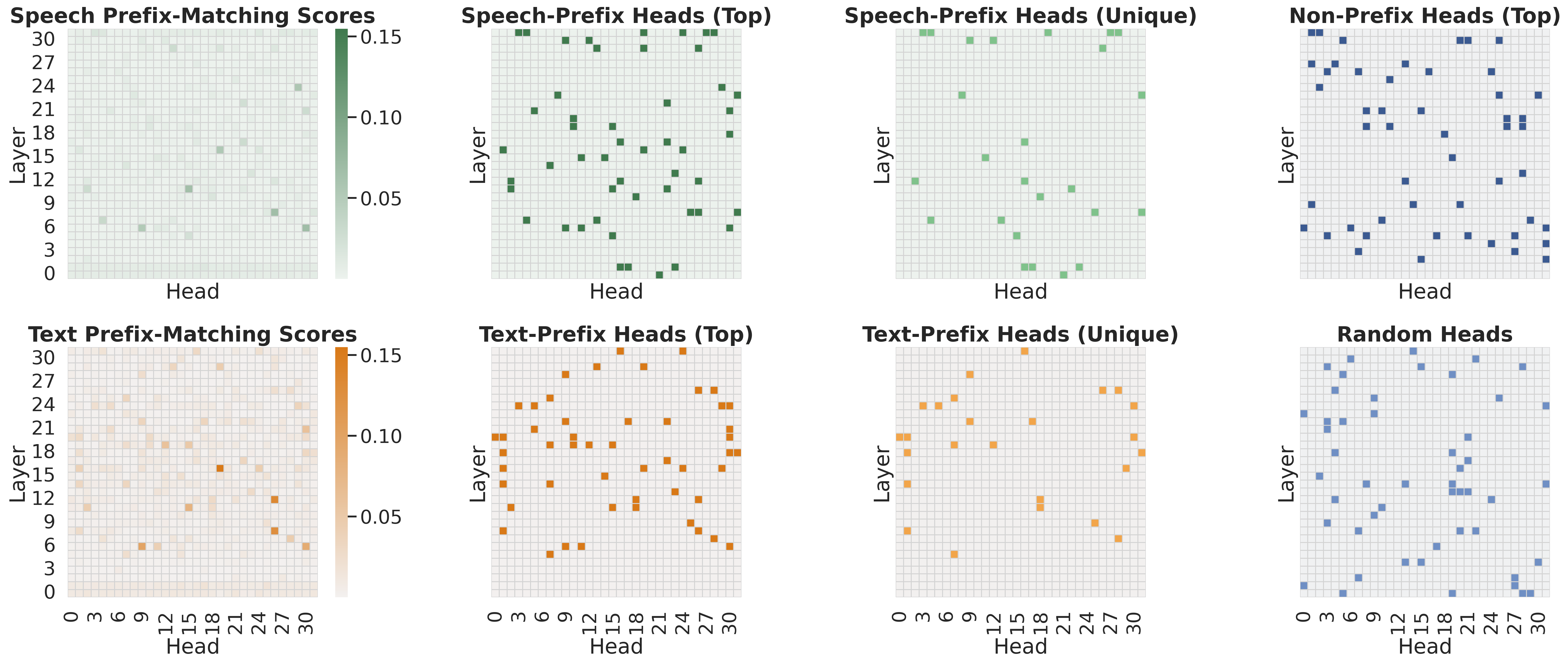}
        \caption{Left: Prefix-matching scores for all heads in SpiritLM, seperately for the speech and text modality. Right: Subsets of heads as specified in \autoref{subsec:head-groups}.}
        \label{fig:heatmaps_head_groups}
    \end{subfigure}

    \vspace{0.5em}

    \begin{subfigure}{\linewidth}
        \centering
        \includegraphics[width=0.8\textwidth]{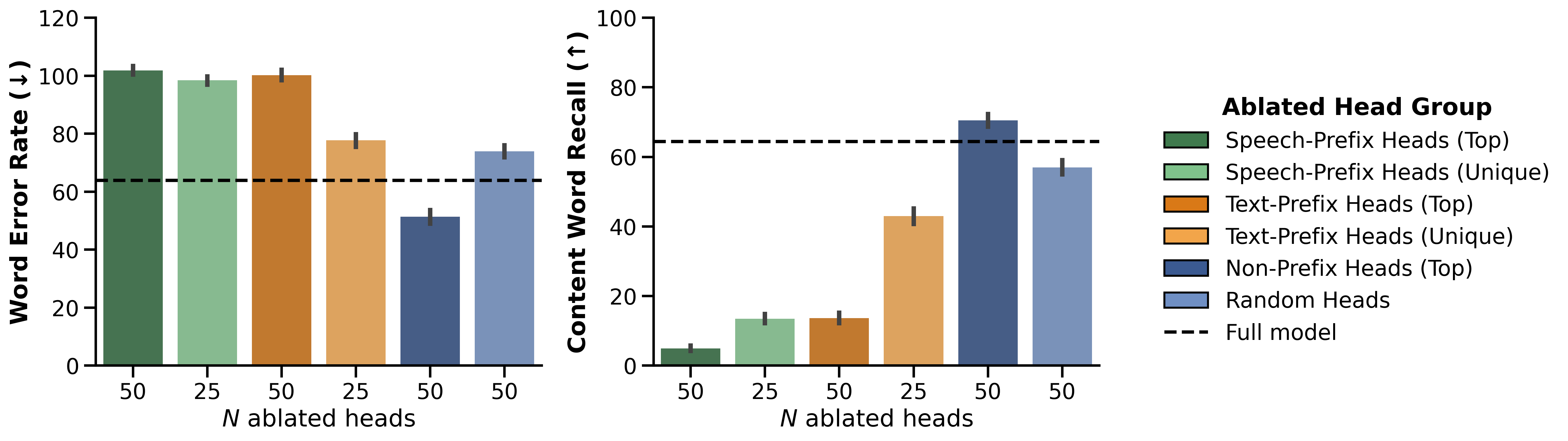}
        \caption{SpiritLM's ICL performance under different ablations. Error bars indicate 95\% confidence intervals.}
        \label{fig:summary_ablated_heads_with_baseline}
    \end{subfigure}

    \caption{Identifying and ablating induction heads in SpiritLM.}
    \label{fig:ablation_results}
\end{figure}

\subsection{Results}

\autoref{fig:heatmaps_head_groups} shows the text and speech prefix-matching scores for all heads in SpiritLM, along with the heads that belong to each head group. We observe substantial overlap between the Speech-Prefix Heads (Top) and the Text-Prefix Heads (Top), but there are also heads which seem to be modality-specific: among the top 50 heads per modality, 25 are unique to speech and 25 are unique to text. This suggests that some heads act as general-purpose prefix matchers (insensitive to modality), whereas others specialize in either speech or text.


\autoref{fig:summary_ablated_heads_with_baseline} shows ICL performance under each ablation. Ablating the Speech-Prefix Heads (Top) and the Text-Prefix Heads (Top) leads to a substantial drop in ICL performance. Interestingly, we observe a similarly large drop when ablating only the unique Speech-Prefix Heads, whereas the effect is less pronounced for the unique Text-Prefix Heads, indicating that prefix-matching on speech is more important for this particular ICL task than prefix-matching on text. Importantly, ablating Random Heads results in only a small decrease in performance compared to the full model; this decrease is much smaller than the effect of ablating Prefix-Matching Heads. Finally, a very interesting observation is that ablating the Non-Prefix Heads (Top) actually slightly improves performance, suggesting that these heads may  harm ICL by attending to irrelevant parts of the context (i.e., non-prefix-matching tokens).

\section{Conclusion \& Discussion}

We presented a study of ICL in a SpeechLM, combining empirical and mechanistic perspectives. Empirically, we analyzed how acoustic and linguistic features of the demonstrations influence ICL. We found that speaking rate both affects task performance and is mimicked in the output, while pitch range and intensity have little influence. Moreover, lexical overlap between demonstration and target improves ICL performance, while semantic similarity has a smaller effect. Mechanistically, we identified modality-insensitive and modality-specific induction heads, and showed that ablating them (particularly those that appear to specialize in speech tokens) leads to a substantial degradation in ICL ability, whereas ablating random heads does not, highlighting their causal role in speech-based ICL.

A natural follow-up question is how these empirical and mechanistic findings relate to each other. We hypothesize that the features which most strongly affect ICL do so because they create more opportunities for prefix-matching by induction heads. For example, when there is lexical overlap between demonstration and target, induction heads can more readily induce a token's continuation by attending to its previous occurrence. Similarly, speaking rate modulates the temporal distribution of tokens and thus affects the scope for prefix-matching: when linguistic content is fixed, a slower speaking rate produces relatively more repeated tokens than a faster speaking rate (see Appendix \autoref{appendix-sec:fast-vs-slow}), which induction heads may exploit. Testing this hypothesis is a promising direction for future work.

Two further directions emerge from our analyses. First, it would be interesting to study prefix-matching behavior across larger spans than single tokens, especially given that the token granularity differs between modalities: speech tokens represent 25-millisecond audio frames, whereas text tokens represent subwords. Second, a systematic comparison of induction head behavior in text-only LMs and SpeechLMs would be valuable. Preliminary results in Appendix~\autoref{appendix-sec:prefix-matching-random} show that Llama-2-7B, the text-only backbone of SpiritLM, achieves higher prefix-matching scores on text-only inputs than SpiritLM does. If induction heads are indeed stronger in text-only LMs, this could help explain why text-only LMs generally outperform SpeechLMs on few-shot language understanding and reasoning tasks.

\section*{Acknowledgments}
This research is funded by the Netherlands Organisation for Scientific Research (NWO) through NWA-ORC grant NWA.1292.19.399 for \textit{InDeep}.

\bibliography{colm2026_conference}

@inproceedings{min-etal-2022-rethinking,
    title = "Rethinking the Role of Demonstrations: What Makes In-Context Learning Work?",
    author = "Min, Sewon  and
      Lyu, Xinxi  and
      Holtzman, Ari  and
      Artetxe, Mikel  and
      Lewis, Mike  and
      Hajishirzi, Hannaneh  and
      Zettlemoyer, Luke",
    editor = "Goldberg, Yoav  and
      Kozareva, Zornitsa  and
      Zhang, Yue",
    booktitle = "Proceedings of the 2022 Conference on Empirical Methods in Natural Language Processing",
    month = dec,
    year = "2022",
    address = "Abu Dhabi, United Arab Emirates",
    publisher = "Association for Computational Linguistics",
    url = "https://aclanthology.org/2022.emnlp-main.759/",
    doi = "10.18653/v1/2022.emnlp-main.759",
    pages = "11048--11064"
}

@article{chan2022data,
  title={Data distributional properties drive emergent in-context learning in transformers},
  author={Chan, Stephanie and Santoro, Adam and Lampinen, Andrew and Wang, Jane and Singh, Aaditya and Richemond, Pierre and McClelland, James and Hill, Felix},
  journal={Advances in neural information processing systems},
  volume={35},
  pages={18878--18891},
  year={2022}
}

@article{olsson2022context,
  title={In-context learning and induction heads},
  author={Olsson, Catherine and Elhage, Nelson and Nanda, Neel and Joseph, Nicholas and DasSarma, Nova and Henighan, Tom and Mann, Ben and Askell, Amanda and Bai, Yuntao and Chen, Anna and others},
  journal={arXiv preprint arXiv:2209.11895},
  year={2022}
}

@inproceedings{crosbie2025induction,
  title={Induction heads as an essential mechanism for pattern matching in in-context learning},
  author={Crosbie, Joy and Shutova, Ekaterina},
  booktitle={Findings of the Association for Computational Linguistics: NAACL 2025},
  pages={5034--5096},
  year={2025}
}

@inproceedings{roll2025context,
  title={In-context learning boosts speech recognition via human-like adaptation to speakers and language varieties},
  author={Roll, Nathan and Graham, Calbert and Tatsumi, Yuka and Nguyen, Kim Tien and Sumner, Meghan and Jurafsky, Dan},
  booktitle={Proceedings of the 2025 Conference on Empirical Methods in Natural Language Processing},
  pages={4412--4426},
  year={2025}
}

@inproceedings{linke2025context,
  title={Context is all you need? Low-resource conversational ASR profits from context, coming from the same or from the other speaker},
  author={Linke, Julian and Winkler, Jana and Schuppler, Barbara},
  booktitle={Proc. Interspeech 2025},
  pages={3199--3203},
  year={2025}
}

@article{touvron2023llama,
  title={Llama: Open and efficient foundation language models},
  author={Touvron, Hugo and Lavril, Thibaut and Izacard, Gautier and Martinet, Xavier and Lachaux, Marie-Anne and Lacroix, Timoth{\'e}e and Rozi{\`e}re, Baptiste and Goyal, Naman and Hambro, Eric and Azhar, Faisal and others},
  journal={arXiv preprint arXiv:2302.13971},
  year={2023}
}

@article{sinclair2022structural,
  title={Structural persistence in language models: Priming as a window into abstract language representations},
  author={Sinclair, Arabella and Jumelet, Jaap and Zuidema, Willem and Fern{\'a}ndez, Raquel},
  journal={Transactions of the Association for Computational Linguistics},
  volume={10},
  pages={1031--1050},
  year={2022},
  publisher={MIT Press One Broadway, 12th Floor, Cambridge, Massachusetts 02142, USA~…}
}

@article{nguyen-etal-2025-spirit,
    title = "{S}pi{R}it-{LM}: Interleaved Spoken and Written Language Model",
    author = "Nguyen, Tu Anh  and
      Muller, Benjamin  and
      Yu, Bokai  and
      Costa-jussa, Marta R.  and
      Elbayad, Maha  and
      Popuri, Sravya  and
      Ropers, Christophe  and
      Duquenne, Paul-Ambroise  and
      Algayres, Robin  and
      Mavlyutov, Ruslan  and
      Gat, Itai  and
      Williamson, Mary  and
      Synnaeve, Gabriel  and
      Pino, Juan  and
      Sagot, Beno{\^i}t  and
      Dupoux, Emmanuel",
    journal = "Transactions of the Association for Computational Linguistics",
    volume = "13",
    year = "2025",
    address = "Cambridge, MA",
    publisher = "MIT Press",
    url = "https://aclanthology.org/2025.tacl-1.2/",
    doi = "10.1162/tacl_a_00728",
    pages = "30--52"
}

@inproceedings{cui-etal-2025-recent,
    title = "Recent Advances in Speech Language Models: A Survey",
    author = "Cui, Wenqian  and
      Yu, Dianzhi  and
      Jiao, Xiaoqi  and
      Meng, Ziqiao  and
      Zhang, Guangyan  and
      Wang, Qichao  and
      Guo, Steven Y.  and
      King, Irwin",
    editor = "Che, Wanxiang  and
      Nabende, Joyce  and
      Shutova, Ekaterina  and
      Pilehvar, Mohammad Taher",
    booktitle = "Proceedings of the 63rd Annual Meeting of the Association for Computational Linguistics (Volume 1: Long Papers)",
    month = jul,
    year = "2025",
    address = "Vienna, Austria",
    publisher = "Association for Computational Linguistics",
    url = "https://aclanthology.org/2025.acl-long.682/",
    doi = "10.18653/v1/2025.acl-long.682",
    pages = "13943--13970",
    ISBN = "979-8-89176-251-0"
}

@misc{radford2022whisper,
  doi = {10.48550/ARXIV.2212.04356},
  url = {https://arxiv.org/abs/2212.04356},
  author = {Radford, Alec and Kim, Jong Wook and Xu, Tao and Brockman, Greg and McLeavey, Christine and Sutskever, Ilya},
  title = {Robust Speech Recognition via Large-Scale Weak Supervision},
  publisher = {arXiv},
  year = {2022},
  copyright = {arXiv.org perpetual, non-exclusive license}
}

@inproceedings{wang2024can,
  title={Can whisper perform speech-based in-context learning?},
  author={Wang, Siyin and Yang, Chao-Han and Wu, Ji and Zhang, Chao},
  booktitle={ICASSP 2024-2024 IEEE International Conference on Acoustics, Speech and Signal Processing (ICASSP)},
  pages={13421--13425},
  year={2024},
  organization={IEEE}
}

@misc{praat,
  author = {Boersma, Paul and Weenink, David},
  title = {Praat: doing phonetics by computer {[Computer program]}},
  url = {https://hadoop.apache.org},
  version = {6.4},
  year = {2023},
  url = {https://www.praat.org},
  urldate = {2023-11-15},
}

@inproceedings{ao2022speecht5,
  title={SpeechT5: Unified-Modal Encoder-Decoder Pre-Training for Spoken Language Processing},
  author={Ao, Junyi and Wang, Rui and Zhou, Long and Wang, Chengyi and Ren, Shuo and Wu, Yu and Liu, Shujie and Ko, Tom and Li, Qing and Zhang, Yu and others},
  booktitle={Proceedings of the 60th Annual Meeting of the Association for Computational Linguistics (Volume 1: Long Papers)},
  pages={5723--5738},
  year={2022}
}

@article{kong2020hifi,
  title={Hifi-gan: Generative adversarial networks for efficient and high fidelity speech synthesis},
  author={Kong, Jungil and Kim, Jaehyeon and Bae, Jaekyoung},
  journal={Advances in neural information processing systems},
  volume={33},
  pages={17022--17033},
  year={2020}
}

@article{hsu2021hubert,
  title={Hubert: Self-supervised speech representation learning by masked prediction of hidden units},
  author={Hsu, Wei-Ning and Bolte, Benjamin and Tsai, Yao-Hung Hubert and Lakhotia, Kushal and Salakhutdinov, Ruslan and Mohamed, Abdelrahman},
  journal={IEEE/ACM transactions on audio, speech, and language processing},
  volume={29},
  pages={3451--3460},
  year={2021},
  publisher={IEEE}
}

@inproceedings{NEURIPS2020_1457c0d6,
 author = {Brown, Tom and Mann, Benjamin and Ryder, Nick and Subbiah, Melanie and Kaplan, Jared D and Dhariwal, Prafulla and Neelakantan, Arvind and Shyam, Pranav and Sastry, Girish and Askell, Amanda and Agarwal, Sandhini and Herbert-Voss, Ariel and Krueger, Gretchen and Henighan, Tom and Child, Rewon and Ramesh, Aditya and Ziegler, Daniel and Wu, Jeffrey and Winter, Clemens and Hesse, Chris and Chen, Mark and Sigler, Eric and Litwin, Mateusz and Gray, Scott and Chess, Benjamin and Clark, Jack and Berner, Christopher and McCandlish, Sam and Radford, Alec and Sutskever, Ilya and Amodei, Dario},
 booktitle = {Advances in Neural Information Processing Systems},
 editor = {H. Larochelle and M. Ranzato and R. Hadsell and M.F. Balcan and H. Lin},
 pages = {1877--1901},
 publisher = {Curran Associates, Inc.},
 title = {Language Models are Few-Shot Learners},
 url = {https://proceedings.neurips.cc/paper_files/paper/2020/file/1457c0d6bfcb4967418bfb8ac142f64a-Paper.pdf},
 volume = {33},
 year = {2020}
}

@article{Ferrando2024APO,
  title={A Primer on the Inner Workings of Transformer-based Language Models},
  author={Javier Ferrando and Gabriele Sarti and Arianna Bisazza and Marta Ruiz Costa-juss{\`a}},
  journal={ArXiv},
  year={2024},
  volume={abs/2405.00208},
  url={https://api.semanticscholar.org/CorpusID:269484740}
}

@inproceedings{
wiegreffe2024mechanistic,
title={Mechanistic?},
author={Sarah Wiegreffe and Naomi Saphra},
booktitle={The 7th BlackboxNLP Workshop},
year={2024},
url={https://openreview.net/forum?id=schAf4BPtD}
}

@inproceedings{von2023transformers,
  title={Transformers learn in-context by gradient descent},
  author={Von Oswald, Johannes and Niklasson, Eyvind and Randazzo, Ettore and Sacramento, Jo{\~a}o and Mordvintsev, Alexander and Zhmoginov, Andrey and Vladymyrov, Max},
  booktitle={International Conference on Machine Learning},
  pages={35151--35174},
  year={2023},
  organization={PMLR}
}

@inproceedings{jumelet2024language,
  title={Do language models exhibit human-like structural priming effects?},
  author={Jumelet, Jaap and Zuidema, Willem and Sinclair, Arabella},
  booktitle={Findings of the Association for Computational Linguistics: ACL 2024},
  pages={14727--14742},
  year={2024}
}

@inproceedings{zhou2025context,
  title={Is in-context learning a type of error-driven learning? evidence from the inverse frequency effect in structural priming},
  author={Zhou, Zhenghao and Frank, Robert and McCoy, R Thomas},
  booktitle={Proceedings of the 2025 Conference of the Nations of the Americas Chapter of the Association for Computational Linguistics: human language technologies (volume 1: long papers)},
  pages={11712--11725},
  year={2025}
}

@inproceedings{liu2022makes,
  title={What makes good in-context examples for GPT-3?},
  author={Liu, Jiachang and Shen, Dinghan and Zhang, Yizhe and Dolan, William B and Carin, Lawrence and Chen, Weizhu},
  booktitle={Proceedings of Deep Learning Inside Out (DeeLIO 2022): The 3rd workshop on knowledge extraction and integration for deep learning architectures},
  pages={100--114},
  year={2022}
}

@inproceedings{zhao2021calibrate,
  title={Calibrate before use: Improving few-shot performance of language models},
  author={Zhao, Zihao and Wallace, Eric and Feng, Shi and Klein, Dan and Singh, Sameer},
  booktitle={International conference on machine learning},
  pages={12697--12706},
  year={2021},
  organization={Pmlr}
}

@inproceedings{bansal-etal-2023-rethinking,
    title = "Rethinking the Role of Scale for In-Context Learning: An Interpretability-based Case Study at 66 Billion Scale",
    author = "Bansal, Hritik  and
      Gopalakrishnan, Karthik  and
      Dingliwal, Saket  and
      Bodapati, Sravan  and
      Kirchhoff, Katrin  and
      Roth, Dan",
    editor = "Rogers, Anna  and
      Boyd-Graber, Jordan  and
      Okazaki, Naoaki",
    booktitle = "Proceedings of the 61st Annual Meeting of the Association for Computational Linguistics (Volume 1: Long Papers)",
    month = jul,
    year = "2023",
    address = "Toronto, Canada",
    publisher = "Association for Computational Linguistics",
    url = "https://aclanthology.org/2023.acl-long.660/",
    doi = "10.18653/v1/2023.acl-long.660",
    pages = "11833--11856"
}

@article{elhage2021mathematical,
   title={A Mathematical Framework for Transformer Circuits},
   author={Elhage, Nelson and Nanda, Neel and Olsson, Catherine and Henighan, Tom and Joseph, Nicholas and Mann, Ben and Askell, Amanda and Bai, Yuntao and Chen, Anna and Conerly, Tom and DasSarma, Nova and Drain, Dawn and Ganguli, Deep and Hatfield-Dodds, Zac and Hernandez, Danny and Jones, Andy and Kernion, Jackson and Lovitt, Liane and Ndousse, Kamal and Amodei, Dario and Brown, Tom and Clark, Jack and Kaplan, Jared and McCandlish, Sam and Olah, Chris},
   year={2021},
   journal={Transformer Circuits Thread},
   note={https://transformer-circuits.pub/2021/framework/index.html}
}

@article{zhang2023makes,
  title={What makes good examples for visual in-context learning?},
  author={Zhang, Yuanhan and Zhou, Kaiyang and Liu, Ziwei},
  journal={Advances in Neural Information Processing Systems},
  volume={36},
  pages={17773--17794},
  year={2023}
}

@article{Chang2006BecomingS,
  title={Becoming syntactic.},
  author={Franklin Chang and Gary S. Dell and Kathryn Bock},
  journal={Psychological review},
  year={2006},
  volume={113 2},
  pages={
          234-72
        },
  url={https://api.semanticscholar.org/CorpusID:1237448}
}

@inproceedings{dong2024survey,
  title={A survey on in-context learning},
  author={Dong, Qingxiu and Li, Lei and Dai, Damai and Zheng, Ce and Ma, Jingyuan and Li, Rui and Xia, Heming and Xu, Jingjing and Wu, Zhiyong and Chang, Baobao and others},
  booktitle={Proceedings of the 2024 conference on empirical methods in natural language processing},
  pages={1107--1128},
  year={2024}
}
\bibliographystyle{colm2026_conference}

\appendix
\section{Appendix}

\subsection{SpiritLM Generation Setup}\label{appendix-sec:generation-setup}

\begin{itemize}
    \item \textbf{Model:}
    \begin{itemize}
        \item \texttt{Spiritlm("spirit-lm-base-7b")}
    \end{itemize}
    \item \textbf{Generation configuration:}
    \begin{itemize}
        \item \texttt{output\_modality = "speech"}
        \item \texttt{max\_new\_tokens = 50}
        \item \texttt{do\_sample = False}
        \item \texttt{seed = 42}
    \end{itemize}
\end{itemize}

\subsection{Performance of Whisper Model Sizes across Speech Types}\label{appendix-sec:whisper-size-comparison}

\begin{figure}[h]
\centering 
\includegraphics[width=0.9\textwidth]{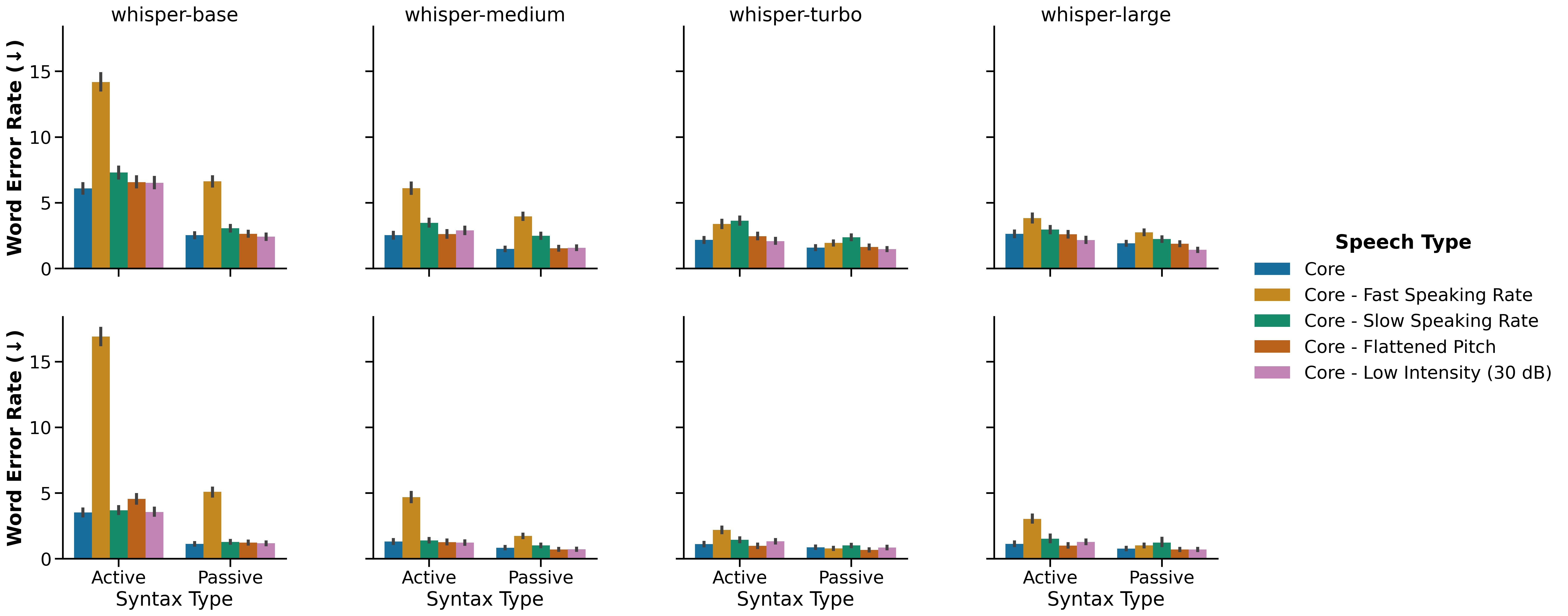}
\caption{Word Error Rate of different Whisper model sizes on various speech types.}
\label{fig:whisper-comparison}
\end{figure}

\autoref{fig:whisper-comparison} shows the performance of different sizes of Whisper across speech types. We evaluated models on the demonstration and target sentences from the \textit{Core} condition of \textsc{Prime-LM}, synthesized and acoustically manipulated in different ways (i.e., the input data for SpiritLM in our ICL experiments, described in \autoref{sec:data}). The top row shows the performance on sentences synthesized by SpeechT5; the bottom row shows performance on sentences synthesized by KokoroTTS. We observe that smaller models (\texttt{base} and \texttt{medium}) perform worse on speech with a \textit{Fast Speaking Rate} compared to other speech types; this pattern disappears for the larger models (\texttt{turbo} and \texttt{large}).

\subsection{Main Results for SpeechT5}\label{appendix-sec:speecht5-results}

\begin{figure}[h]
\centering 
\includegraphics[width=0.8\textwidth]{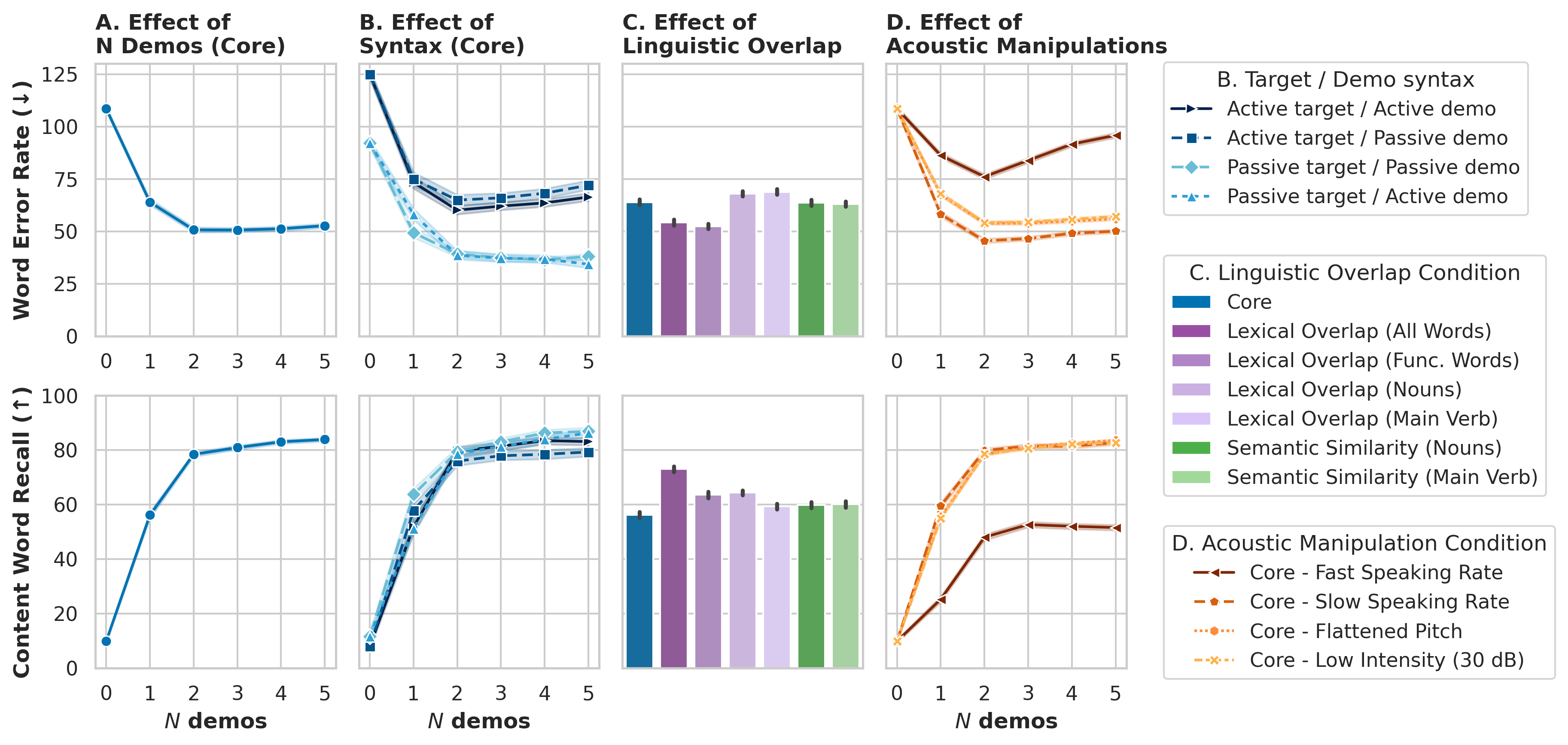}
\caption{ICL performance of SpiritLM given demonstrations synthesized by SpeechT5. \textbf{(A)}: performance on the \textit{Core} condition across 1-5 demonstrations; \textbf{(B)}: performance on the \textit{Core} condition, separated by the syntactic structure of the target and the demonstration; \textbf{(C)}: performance under various types of linguistic overlap between demonstration and target; \textbf{(D)}: performance under various acoustic manipulations of the demonstrations. Shaded regions/error bars indicate 95\% confidence intervals.}
\label{fig:ling_manip_speecht5}
\end{figure}

In \autoref{fig:ling_manip_speecht5}, we show the results from \autoref{fig:ling_manip_kokoro}, but with demonstrations synthesized by SpeechT5.

\subsection{Prefix-Matching on Random Sequences}\label{appendix-sec:prefix-matching-random}

We follow the behavioral method by \cite{olsson2022context} (and used in \cite{bansal-etal-2023-rethinking} and \cite{crosbie2025induction}) for computing prefix-matching scores as a proxy for mechanistically identifying induction heads. Concretely, we generate a sequence of 50 unique random tokens, excluding the top 4\% most frequent and the bottom 4\% least frequent tokens in the vocabulary. We then repeat this 50-token sequence four times and feed the resulting sequence to the model.

For each attention head, we compute the prefix-matching score as follows: for each token $t$ at position $i$, we look up $t_i$ in an earlier repeat (we refer to this token as $t_j$ at position $j < i$). We then include the attention from $t_i$ to $t_{j+1}$ (i.e., the token immediately following $t_j$) in the \emph{prefix-matching} attention sum, and average this sum across the number of tokens in repeats 1, 2, and 3 (i.e., 150 tokens; repeat 0 is not included as this repeat does not have previous context). We repeat this procedure for $N$ random sequences and take the average prefix-matching score across sequences. Whereas previous work used $N = 5$, we use $N = 25$ for additional robustness.

Because SpiritLM operates on both speech and text tokens, we compute prefix-matching scores on three types of random sequences: \textbf{(1) Text-only:} we sample 50 tokens from the text vocabulary only; \textbf{(2) Speech-only:} we sample 50 tokens from the speech vocabulary only; \textbf{(3) Interleaved:} we sample 25 tokens from the text vocabulary and 25 from the speech vocabulary and interleave them in random order.

\begin{figure}[h]
\centering 
\includegraphics[width=\textwidth]{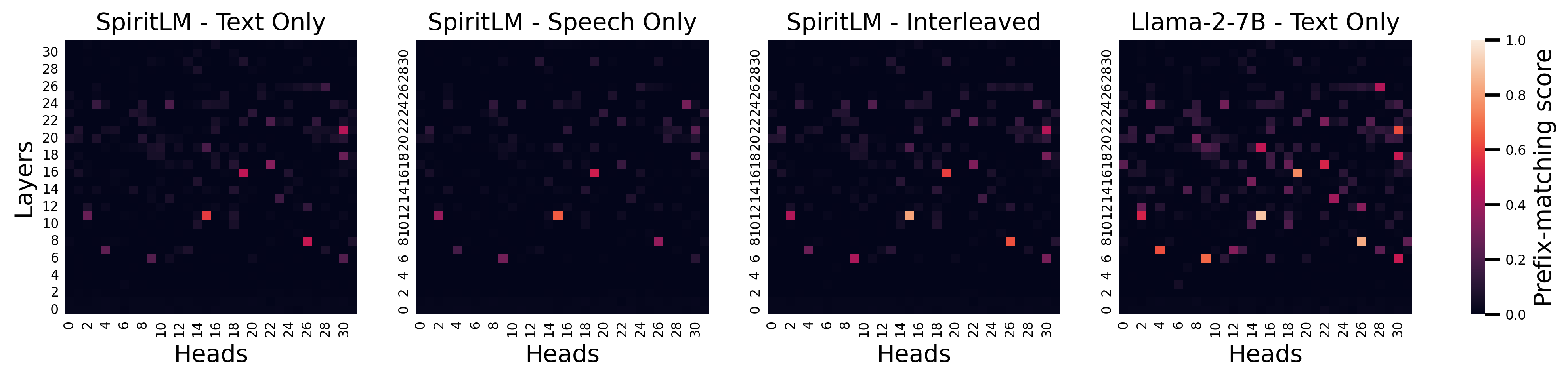}
\caption{Prefix-matching scores for each head in SpiritLM on sequences of random tokens (text tokens only, speech tokens only, or interleaved text and speech tokens). The results of Llama-2-7B (the text-only backbone of SpiritLM) are added as a reference.}
\label{fig:heatmaps-prefix-matching}
\end{figure}

\paragraph{Results.} \autoref{fig:heatmaps-prefix-matching} shows the prefix-matching results on random sequences. We observe a similar distribution of prefix-matching heads across all three input types, although the strength of prefix-matching scores varies depending on the input.

For comparison, we include the prefix-matching scores of Llama-2-7B, i.e., the text-only backbone of SpiritLM. Interestingly, the scores are generally higher in the Llama model, indicating that continuously pretraining this model on speech may have reduced induction behavior --- this observation should be verified in future work.

\subsection{Prefix-Matching on Slow versus Fast Speech}\label{appendix-sec:fast-vs-slow}

\begin{figure}[h]
\centering 
\includegraphics[width=0.9\textwidth]{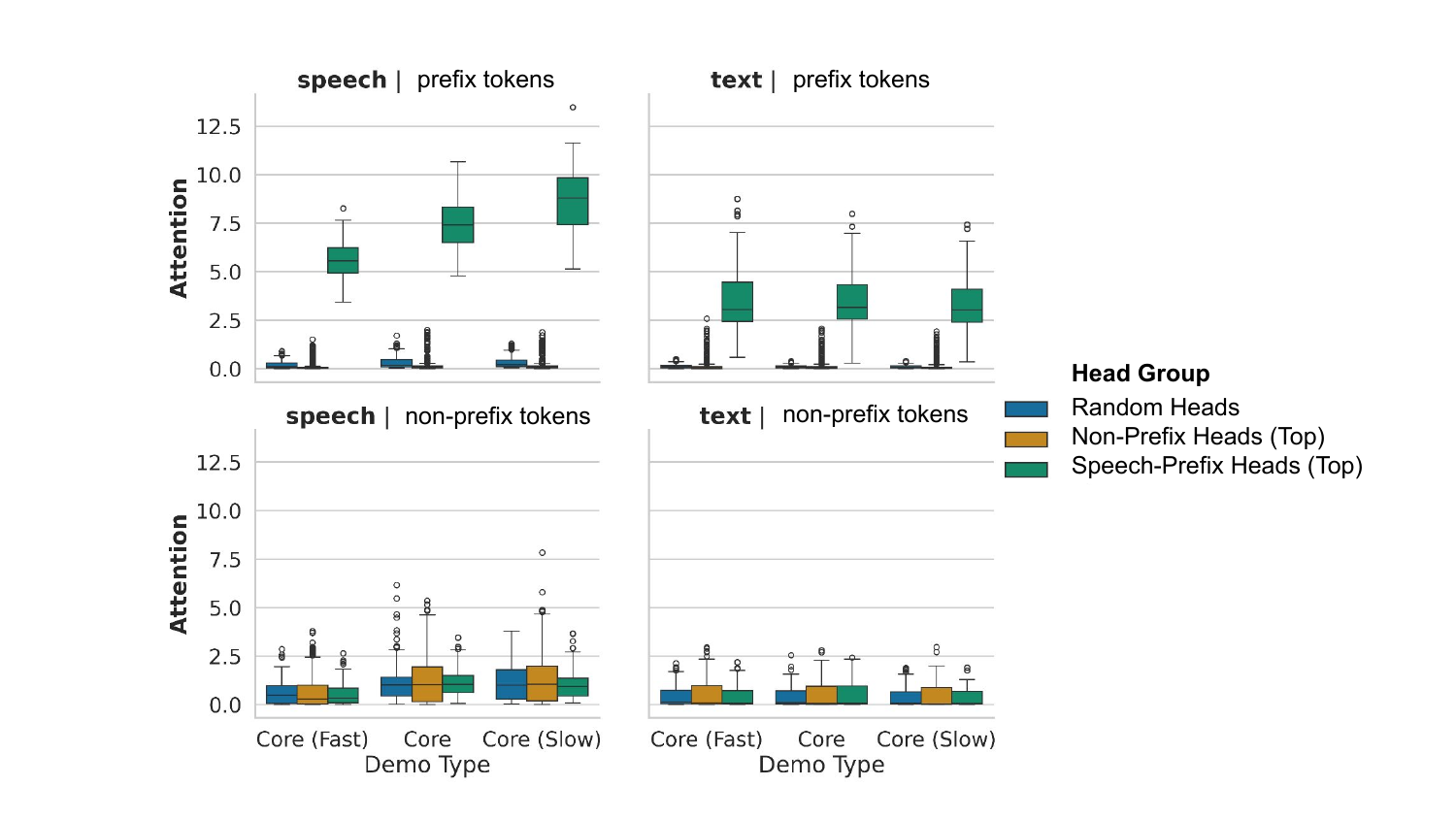}
\caption{Attention to prefix-matching and non-prefix-matching tokens (speech vs. text) by different head groups, across three different demonstration types.}
\label{fig:fast-vs-slow}
\end{figure}

\autoref{fig:fast-vs-slow} shows the attention allocated to prefix-matching and non-prefix-matching tokens (speech vs. text) by three groups of heads: \textit{Speech-Prefix Heads (Top)}, \textit{Non-Prefix Heads (Top)}, and \textit{Random Heads}. Results are shown separately for the \textit{Core}, \textit{Core – Fast Speaking Rate}, and \textit{Core – Slow Speaking Rate} conditions. The plotted values correspond to the scores defined in \autoref{sec:pfx-matching-icl-prompts}, prior to normalization by the number of tokens. For the Speech-Prefix Heads (Top), we observe that the attention on speech prefix-matching tokens is relatively higher in the \textit{Slow} condition compared to the \textit{Fast} condition, which may be beneficial for ICL.

\end{document}